# MODELING AND SIMULATING RETAIL MANAGEMENT PRACTICES: A FIRST APPROACH

**ABSTRACT**

Multi-agent systems offer a new and exciting way of understanding the world of work. We apply agent-based modeling and simulation to investigate a set of problems in a retail context. Specifically, we are working to understand the relationship between people management practices on the shop-floor and retail performance. Despite the fact we are working within a relatively novel and complex domain, it is clear that using an agent-based approach offers great potential for improving organizational capabilities in the future.

Our multi-disciplinary research team has worked closely with one of the UK's top ten retailers to collect data and build an understanding of shop-floor operations and the key actors in a department (customers, staff, and managers). Based on this case study we have built and tested our first version of a retail branch agent-based simulation model where we have focused on how we can simulate the effects of people management practices on customer satisfaction and sales. In our experiments we have looked at employee development and cashier empowerment as two examples of shop-floor management practices.

In this paper we describe the underlying conceptual ideas and the features of our simulation model. We present a selection of experiments we have conducted in order to validate our simulation model and to show its potential for answering "what-if" questions in a retail context. We also introduce a novel performance measure which we have created to quantify customers' satisfaction with service, based on their individual shopping experiences.

## 1  BACKGROUND

### 1.1  Why is Retail Productivity Important?

The retail sector significantly contributes to the UK's relatively low productivity compared to France, Germany and the USA (Reynolds et al., 2005), popularly termed the 'productivity gap'. A large-scale literature review of management practices and retail performance and productivity (Siebers et al., 2008) concluded that management practices are multidimensional constructs which tend to demonstrate a complex relationship with productivity. The authors concluded that it may be the context-specific nature of management practices and productivity which precludes clear patterns in the results of empirical studies (for a further review see Wall & Wood 2005). Many experts agree that the focus is shifting to looking inside organizations to understand the source of the problem (Delbridge et al., 2006). Focusing on management practices may offer an opportunity to further our understanding of the UK's relatively low levels of retail productivity (Porter & Ketels, 2003).

## 1.2 Customer Satisfaction and In-Store Experiences

Without customers a retailer is without a business; Hill and Alexander (2006, p.11) advocate the only route to success is, "Do best what matters to customers." Measuring customer satisfaction is the key way in which a retailer can quantify and understand their strengths and weaknesses. Empirical evidence suggests there is a need to differentiate between the components of a global customer satisfaction measure (Rust et al., 1995; Garbarino & Johnson, 1999). Consequently, instead of focusing on overall satisfaction as a global evaluation, we will investigate *customer satisfaction* as it is empirically driven by visitors' in-store experiences and perceptions of service they receive (Torres et al., 2001).

## 1.3 People Management Practices

Managers working in retail stores tend to be under a lot of pressure to allocate their time effectively and to prioritize the competing tasks and impromptu issues that can crop up throughout their shift. People management practices offer a way of enhancing the overall operation of the store through its staff. People (or Human Resource) management practices have been defined as the, "… organizational activities directed at managing the pool of human capital and ensuring that the capital is employed towards the fulfillment of organizational goals," (p.304, Wright et al., 1994). Examples of people management practices are empowerment, team-based working and skill development.

## 2 INTRODUCTION

There exists a large body of work investigating the modeling and simulation of operational management practices, whereas people management practices have often been neglected. Yet research suggests that people management practices crucially impact upon an organization's performance (for example, Birdi et al., 2008).

The overall aim of our project is to understand and predict the impact of different people management practices on retail productivity and performance. One key objective has been to apply simulation to devise a functional representation of the retail shop-floor driven by a real system. To achieve this objective we have adopted a case study approach and integrated applied research methods to collect complementary qualitative and quantitative data. In summary, we have conducted four weeks of informal participant observation, forty staff interviews supplemented by a short questionnaire regarding the effectiveness of various management practices, and drawn upon a range of internal company documentation. Early experimentation with the model has led us to develop and advance its operation in order to facilitate more comprehensive investigation of the impact of management practices. By reducing the level of abstraction within the model we are able to evaluate simulation runtime outcomes in terms more closely linked to those of the real system.

In this paper we describe the development of the first version of our simulation model where we have focused mainly on how we can simulate the effects of people management practices on customer satisfaction and sales. We have chosen employee development and empowerment as examples of such people management practices. In Sections 3, 4, and 5 we provide an overview of the research we have completed leading up to the creation of our simulation model of two retail departments. Section 3 embeds our selection of modeling technique in the broader modeling literature, and Section 4 describes the model design including data collection, model conceptualization, and a description of how we incorporate the empirical data we have gathered during our case study work. Section 5 explains the implementation of these concepts and the data according to the first full version of our simulation model (referred to as ManPraSim v1). In Section 6 we present two sets of validation experiments, and then three sets of operational experiments to investigate to impact of management practices on department performance measures including customers' satisfaction with the service provided. We draw some conclusions in Section 7 and identify priorities for future work.

## 3      WHY USE AGENT-BASED SIMULATION?

There are a number of competing approaches to modeling, and the decision to choose Agent-Based Modeling and Simulation (ABMS) followed a careful review and evaluation of different approaches.

### 3.1    Selection of modeling technique

Operations Research (OR) is applied to problems concerning the conduct and co-ordination of the operations within an organization (Hillier & Lieberman, 2005). An OR study usually involves the development of a scientific model which attempts to abstract the essence of the real problem. When investigating the behavior of a complex system it is very important to select an appropriate modeling technique. In order to be able to make a choice for our project, we reviewed the relevant literature spanning the fields of Economics, Organizational Behavior, Psychology, Retail, Marketing, OR, Artificial Intelligence, and Computer Science. Within these fields a wide variety of modeling approaches are used which can be classified into three main categories: analytical approaches, heuristic approaches, and simulation. In many cases we found modelers had adopted an integrated approach and applied more than one technique within a single model. Common combinations were 'simulation / analytical' for comparing efficiency of alternative future scenarios (e.g. Greasley, 2005), and 'simulation / analytical' or 'simulation / heuristic' where analytical or heuristic models were used to represent the behavior of the entities within the simulation model (e.g. Schwaiger & Stahmer, 2003).

Simulation opens the door to a new way of thinking about social and economic processes, based on ideas about the emergence of complex behavior from relatively simple activities (Simon, 1996). Whereas analytical models tend to aim to explain correlations between variables measured at one single point in time, simulation models are concerned with the development of a system over time. Furthermore, analytical models usually work on a much higher level of abstraction than simulation models. For simulation models it is critical to define the right level of abstraction.

Csik (2003) states that on the one hand the number of free parameters must be kept as small as possible. On the other hand, too much abstraction and simplification will threaten the ability of the model to accurately represent the real system. OR usually employs three different types of simulation modeling to help understand the behavior of organizational systems, each of which has its distinct application area: Discrete Event Simulation (DES), System Dynamics Simulation (SDS) and Agent Based Simulation (ABS). The choice of the most suitable approach will always depend on the focus of the model, which input data is available, the level of analysis and what kind of answers are sought.

In our review we put particular emphasis on those publications that try to model the link between management practices and productivity or performance in the retail sector. We found a very limited number of papers that investigate management practices in retail at the organizational level, with the majority of these papers focusing on marketing practices (e.g. Keh et al., 2006). Agent-Based Modeling (ABM), using simulation as the method of execution, was by far the most popular technique. It seems to be accepted as the natural way of system representation for organizations; active entities in the live environment are interpreted as actors in the model.

## 3.2   Agent-Based Modeling and Simulation

Although computer simulation has been used widely since the 1960s, ABM only became popular at the start of the 1990s (Epstein & Axtell, 1996). ABM can be used to study how micro-level processes affect macro-level outcomes. A complex system is represented by a collection of individual agents which are programmed to follow simple behavioral rules. Agents can interact with one another and with their environment, and these interactions can result in complex collective behavioral patterns. Macro behavior is not explicitly simulated; it emerges from the micro-decisions and actions of individual agents (Pourdehnad et al., 2002). The main characteristics of agents are: autonomous operation, the ability to act flexibly in response to the environment, and pro-activeness driven by internal motivations. Agents are designed to mimic the attributes and behaviors of their real-world counterparts. Simulation output can be used for explanatory, exploratory and predictive purposes (Twomey & Cadman, 2002).

The way in which agents are modeled appears to be more suitable than DES for modeling human-centric complex adaptive systems (Siebers, 2006). There is a structural correspondence between the real system and the model representation, which makes these models more intuitive and easier to understand than for example a system of differential equations as used in SDS. Hood (1998) emphasized one of the key strengths of this technique is that the system as a whole is not constrained to exhibit any particular behavior because the system properties emerge from its constituent agent interactions. Consequently assumptions of linearity, equilibrium and so on, are not needed. Of course there are disadvantages; there is a general consensus in the literature that it is difficult to empirically evaluate agent-based models, in particular at the macro level, because the behavior of the system emerges from the interactions between the individual entities (Moss & Edmonds, 2005). Furthermore Twomey & Cadman (2002) state that problems often occur through the lack of adequate empirical data, and that there is always a danger that people new to ABM may expect too much from the models, in particular with respect to predictive ability, though this last criticism applies to all the simulation approaches mentioned above.

Overall we can conclude that ABMS is the most appropriate technique to investigate people management practices. This approach provides us with the opportunity to model organizational characters and their interactions in realistic and valid ways.

## 4   MODEL DESIGN

We emphasize the central role of data collection and understanding of the real system to inform the conceptualization and implementation of our model.

### 4.1   Knowledge gathering

The case studies were conducted in the same two departments across two branches of a leading UK department store. We adopted an integrated approach using a complementary set of data collection methods: participant observation; semi-structured interviews; completion of a management practices questionnaire with team members, managers and personnel managers; and the analysis of company data and reports. Research findings were consolidated and fed back (via report and presentation) to employees and managers with extensive experience and knowledge of the case study departments in order to cross-validate our understanding and conclusions.

Preliminary case study findings suggested that we needed to configure the model to represent the different department types: Audio and Television (A&TV) and Womenswear (WW). This approach also helps to ensure that the simulation results remain as broadly applicable as possible. Case study work revealed substantial differences between the two department types, a divergence which is generally driven by fundamentally different product characteristics. For example, the average purchase in A&TV is more expensive than in WW. The likelihood of a customer seeking advice is higher in A&TV, and the average customer service time is longer than in WW. Customers in WW are more likely to make a purchase than in A&TV.

Our empirical approach to understanding the real case studies has played a crucial role in facilitating the conceptualization of how the real system is structured. This is an important stage to any simulation project, revealing insights into the operation of the system as well as the behavior of and interactions between the different characters in the system.

### 4.2   Conceptual Modeling

To make the most of the empirical data and insights obtained through the case studies, the core aspects of the model were conceptualized and mapped out prior to implementation.

#### 4.2.1   Modeling Approach

Building on the findings from our literature review we have used the agent paradigm to conceptualize and model the actors within the system under investigation.

Our modeling approach has been iterative, firstly creating a relatively simple model and progressively building in more and more complexity. We started by trying to understand the particular problem domain and to generate the underlying rules currently in place. We have since progressed to the process of building an ABS model of the real system using the information gathered during our case study and to validate our model by simulating the operation of the real system.

This approach allows us to assess the accuracy of the system representation. When the simulation has provided a sufficiently good representation we have been able to move to the next stage, and generate new scenarios for how the system could work using new rules.

### 4.2.2  Concept for the Simulation Model

Our conceptual ideas for the simulation model are shown in Figure 1. Within our simulation model we have three different types of agents (customers, shop-floor staff, and managers) each with a different set of relevant attributes. Global parameters can influence any aspect of the system, and define, for example, the number of agents in the system. With regards to system outputs we hope to find some unforeseeable, emergent behavior on the macro level. Maintaining a visual representation of the simulated system and its actors will allow us to closely monitor and better understand the interactions of entities within the system. In addition, by measuring the performance of the system we will be able identify bottlenecks in the real system and to subsequently to optimize it.

[INSERT FIGURE 1]

### 4.2.3  Concept for the Actors

The agents have been designed and represented using state charts. State charts display the different states which an entity can be in and define the transitional events which are the triggers driving an actor's change from one behavioral state to another. This is exactly the information we need in order to represent our agents later within the simulation environment. Furthermore this form of graphical representation is helpful for validating the agent design (micro-level face validation) because it is easy for experts in the real system to understand.

The art of modeling relies on simplification and abstraction (Shannon, 1975). A model is always a restricted copy of the real world and we have to identify the most important components of a system to build effective models. In our case, instead of looking for components we have identified the most important behaviors of each actor and the triggers which initiate a move from one state to another. We have developed state charts for all the relevant actors in our retail department model. Figure 2 shows as an example the conceptual template of a customer agent. The transition rules have been omitted here to keep the chart succinct (see Section 5.1 and 5.2 for a more detailed explanation of the transition rules).

[INSERT FIGURE 2]

Once a customer enters the department he or she will be in the contemplating state. This is a dummy state and represents the reality of an individual thinking through their behavioral intentions prior to acting (Ajzen, 1985), regardless of whether the department visit will result in a planned or unanticipated purchase (Kelly et al., 2000). Even when a particular purchase is planned, the customer may change their mind and go for a substitute product, if they buy at all. He or she will probably start browsing and after a certain amount of time, he or she may require help, queue at the till or leave the shop. If the customer requires help, he or she considers what to do and seeks help by looking for a staff member and will either immediately receive help or wait for attention. If no staff member is available, he or she has to join a queue and wait for help. If the queue moves very slowly it could result in a customer becoming fed up of waiting or running out of time and so he or she leaves the queue prematurely. This does not mean necessarily that he or she will not make a purchase. Sometimes customers would still make a purchase even without getting the advice they were seeking. Another reason why a customer might come into the department is to ask for a refund. We have added this activity to the conceptual model because we will later experiment with different refund policies. From an organizational point of view the refund process is very similar to the help process. The difference is that the refund process will take place at the till. After the refund process is concluded the customer will either continue shopping (i.e. start browsing) or leave the department.

It is important to observe that there is a sequential order to these events which is incorporated into the customer state chart. Furthermore, there is a logical flow to these states. Thus, for example, a customer is unlikely to be queuing at the till in WW to buy something without having first picked up an item. Therefore, the condition for queuing at the till to buy something would be that the customer has been browsing before to pick up an item. These rules have been considered in the implementation (see Section 5.1 and Figure 3 for more details).

During the process of conceiving the model we have questioned whether or not our agents are intelligent. Wooldridge (2002) stated that in order to be intelligent, agents need to be reactive, proactive and social. This is a widely accepted view. Being reactive means responding to changes in the environment (in a timely manner), being proactive means persistently pursuing goals and being social means interacting with other agents (Padgham & Winikoff, 2004). Our agents perceive a goal in that they intend to either make a purchase or return a previous purchase. The buying process has a sub goal; the customer is trying to buy the right thing. If the customer is not sure he or she may ask for help from a shop floor worker. Our agents are not only reactive but also flexible, i.e. they are capable to recover from a failure of action. They have alternatives inbuilt when they are unable to realize their goal in a timely manner. For example if a customer wants to pay but the queue is not moving he or she will always have the chance to leave a queue and pursue another action. This example illustrates that customers can respond in a flexible way to certain changes in their environment, in this case the length of the queue. Finally, as there is communication between agents and staff, they can also be regarded as social entities interacting with others[1].

---

[1] An extensive discussion of the notion of 'intelligence', a topic which seems to split the simulation community (encompassing the ABM community) into two halves, can be found in SIMSOC (2008).

### 4.2.4 Concept for a Customer Satisfaction Measure

Customer perceptions are crucial to measuring the impact of retail management practices. Applied in conjunction with objective performance measures (e.g. sales turnover), it becomes possible to obtain a rounded view of retail performance. Customer service is by definition intangible, and an index of customer satisfaction offers an invaluable way of quantifying customers' perceptions of this. Measures of customer satisfaction are important to provide an indicator of not only how the business is performing at present, but also an idea of how many recent customers will return to the retailer.

Global customer satisfaction is a multi-dimensional construct, an accumulation of separate satisfaction evaluations of multiple facets (see for example Parasuraman et al., 1988; Bolton & Drew, 1991; Mihelis et al., 2001). We conceptualize a measure which draws upon a subset of these facets; focusing on those aspects of satisfaction which we observed to be most pertinent to customer satisfaction in a retail department. We will use customers' perceptions of their in-store experiences, in particular the service that is provided, as an indicator of customer satisfaction. The aim is to go beyond the capabilities of existing measures to create a dynamic measure that considers each step of the entire shopping experience as each individual customer perceives it. The link between shop-floor management practices and this measure is salient because achieving a high level of customer satisfaction is hinges on the availability of suitably skilled staff when customers need them. Relating these to Mihelis and colleagues' (2001) model of global satisfaction, for example, these components relate to two of five high-level components: service (e.g. waiting times, service processes) and personnel (e.g. skills and knowledge).

Previous work examining service encounters in retail settings has shown that the attitudes and behaviors of employees can positively influence customers' perceptions of quality, satisfaction, and hence purchase intentions (Babin et al., 1999; Baker et al., 2002; Dabholkar et al., 1995; Parasuraman et al., 1994). Further to this, key aspects of customer service quality have been shown to impact positively on customer perceptions, and these include circumstances when employees have been perceived as respectful, friendly, knowledgeable about products, responsive to the customer's needs and questions, able to give advice, and have not pursued a 'hard sell' (Darian et al., 2001; Leo & Philippe, 2002). Some businesses continue to gain competitive advantage through priding themselves on exceptional customer service. The importance of providing a high quality service to customers is widely accepted as a crucial topic for management success, as demonstrated by dedicated journals, such as 'Managing Service Quality'. A recent large-scale consumer satisfaction study (conducted by Which, cited by Fluke, 2008) surveyed more than 10,000 people and found that, "shoppers are increasingly willing to spend extra for better service." This is convincing evidence that retailers who strive for high levels of customer satisfaction through a favorable in-store experience are reaping the benefits.

Many methods of calculating customer satisfaction sample only those people who visit the store and leave with a purchase. These methods ignore the store visitors who could have made a purchase; in other words data is not collected from unrealized customers. We would argue that the satisfaction of all store visitors is important and valid for the long-term development and survival of the business. Everyone who visits the store will remember their experiences, whether they have been positive or negative, and this could influence his or her decision to come again or whether to make a future purchase (e.g. Meyer, 2008). For this reason our customer satisfaction measure draws on the perceptions of all department visitors, rather than restricting this measure only to those individuals who make a purchase.

## 5  MODEL IMPLEMENTATION

Our simulation has been implemented in AnyLogic™ version 5.5 which is a Java™ based multi-paradigm simulation software (XJ Technologies, 2007). The simulation model is initialized from an Excel™ spreadsheet. We have implemented the knowledge, experience and data obtained from the case studies, resulting in a model which supports the simulation of the two types of departments (A&TV and WW) within which we conducted our case study work.

### 5.1  Implementing the Concept

The simulation model can represent the following actors: customers, service staff (including cashiers, selling staff of two different training levels) and managers. Figure 3 shows a screenshot of the customer and staff agent logic as it has been implemented in AnyLogic™. Boxes represent customer states, arrows transitions, circles with B branches (decision nodes) and numbers denote satisfaction weights which as a whole form the *service level index*.

[INSERT FIGURE 3]

There are two different types of customer goals implemented: making a purchase or obtaining a refund. If a refund is granted, the customer's goal may then change to making a new purchase, or alternatively they will leave the shop straight away. The customer agent template consists of three main blocks which all use a very similar logic. These blocks are 'Help', 'Pay' and 'Refund'. In each block, in the first instance, customers will try to obtain service directly and if they cannot obtain it (no suitable staff member available) they will have to queue. They will then either be served as soon as the right staff member becomes available or they will leave the queue if they do not want to wait any longer (an autonomous decision). A complex queuing system has been implemented to support different queuing rules. In comparison to the customer agent template, the staff agent template is relatively simple. Whenever a customer requests a service and the staff member is available and has the right level of expertise for the task requested, the staff member commences this activity until the customer releases the staff member. While the customer is the active component of the simulation model the staff member is currently passive, simply reacting to requests from the customer. In future we planned to add a more pro-active role for the staff members, for example offering services to browsing customers.

### 5.2  Input Parameters

We have used frequency distributions and probabilities to assign different values to each individual agent. In this way a population is created that reflects the variations in attitudes and behaviors of their real human counterparts. Often agents are based on analytical models or heuristics and in the absences of adequate empirical data theoretical models are employed; we have incorporated data from the real system wherever possible.

We have used frequency distributions for modeling delays between state changes, specifically triangular distributions supplying the time that an event lasts, using the minimum, mode, and maximum duration. Our triangular distributions are based on our own observation and expert estimates in the absence of numerical data. We have collected this information from the two branches and calculated an average value for each department type, building one set of data for A&TV and one set for WW. Table 1 lists some sample frequency distributions that we have used for modeling the A&TV department (the values presented here have been slightly amended to comply with confidentiality restrictions). The distributions have been used as exit rules for most of the states. All remaining exit rules are based on queue development, i.e. the availability of staff.

[INSERT TABLE 1]

We have used probabilities to model the decision making processes. The probabilities are partly based on company data or published data (e.g. conversion rates, that is the percentage of customers who buy something) and where empirical data has not been available we have collected estimates from knowledgeable individuals working in the case study departments (e.g. the patience of a customer before prematurely leaving a queue). Some examples for the probabilities we have used to model the A&TV department can be found in Table 2, and as before we have calculated average values for each department type. The probabilities link to most of the transition rules at the branches where decisions are made about what action to take (e.g. decision to seek help). The remaining decisions are based on the state of the environment (e.g. leaving the queue, if the queue does not get shorter quickly enough).

[INSERT TABLE 2]

### 5.3 Performance Measures

We have built a number of performance measures into the system to help us understand the outcomes of a simulation run. In this paper we look at the number of transactions, staff utilization indices, the number of satisfied customers, and overall customer satisfaction. These measures are defined as follows. The *number of transactions* acts as a proxy for departmental sales turnover, and allows us to draw links between experimental results and the tangible financial outcomes of the real system. The *staff utilization* indices are presented by staff type, either normal or expert, and help us to understand whether the staff team's composition is effectively meeting the demands placed on it by customers. Satisfaction measures have been introduced to allow the satisfaction of customers with their in-store experiences, and any service with which they have been provided, to be recorded throughout the simulated lifetime. The *number of satisfied customers* is the count of customers who have left the department with a positive service level index (i.e. the count of satisfied customers). *Overall customer satisfaction* is the sum of all customers' service level indices when they leave the department (i.e. the sum of all customers' individual satisfaction levels).

To implement customers' perceptions in the model, we have decomposed each customer's shopping experience across different behavioral states and transitions between them (see Figure 3). The *satisfaction weights* define the relative contribution of each customer transition to the satisfaction score; actual figures are notional. The weightings have been allocated in line with empirical data, or where this has not been possible they have been allocated in an intuitive manner. For example, a large evidence base supports the contention that waiting for service can lead to a customer forming a negative impression of the service being received (Bitner et al., 1990; Katz et al., 1991; Taylor, 1994). Figure 3 displays how this finding has been incorporated into the model. If a customer seeks help and locates help immediately his or her satisfaction score is increased by 2 + 2 = 4. If the customer has to wait for help but gets the help in the end, the overall impact on his or her satisfaction cancels out: 2 – 2 = 0. If, however, the customer gets fed up with waiting for help and leaves the queue prematurely, there is a strong adverse impact on his or her satisfaction score: -4. If this customer then leaves the department without buying anything an additional -2 is added to the satisfaction score, so that this customer at the end of his or her shopping trip would contribute zero to the number of satisfied customers count and -6 to the overall customer satisfaction measure.

Implementing satisfaction weights allows us to account for the differential impact of different components of customers' in-store experiences, and build a more realistic measure of customers' satisfaction with their visit. In line with the empirical findings of Westbrook (1981), a simple linear additive model has been followed, whereby a customer's individual satisfaction weights collected at each relevant transition can be summed up to calculate an overall level of satisfaction at the end of a department visit. We measure customers' service satisfaction in two different ways derived from these weightings; number of satisfied customers and overall customer satisfaction. Applied in conjunction with an ABMS approach, we expect to observe interactions with individual customer differences; variations which have been empirically linked to differences in customers' service satisfaction. This helps the analyst to find out to what extent customers underwent a positive or negative shopping experience. It also allows the analyst to put emphasis on different operational aspects and try out the impact of different management strategies.

Individual differences between customers have already been built into the model and there is some potential to extend the modeled variability between customers by introducing heterogeneous customer types (as discussed in Section 7), and so the satisfaction weights remain static (unless the weights themselves are the experimental variable – see Section 6.3). The rationale for modeling the weights in this way is because although it is likely that any single situation will inevitably be perceived in different ways across individuals, it can also be argued that multiple responses will tend to a normal distribution, resulting in a single 'most likely' or mean response. It is the estimated 'most likely' response which has been implemented in the simulation model. Using static satisfaction weights in this way ensures that we can incorporate a dynamic measure of customers' service satisfaction without introducing unnecessary variability.

At the end of the simulation run, there are a certain number of customers who are satisfied, those who are neutral in opinion, and those who are dissatisfied. A higher count of the number of satisfied customers means that more customers are satisfied when they leave the store. A higher (or lower) level of overall customer satisfaction is likely to be the result of a combination of a higher number of customers holding that satisfied (or dissatisfied) point of view and also possessing a more extreme opinion.

## 6  EXPERIMENTING WITH THE MODEL

In this section we present and describe the results from a series of experiments. Firstly we validate the model by varying the operational staffing configuration and examining the impact on sales figures and customers' service satisfaction. Then we test the impact of management practices (task empowerment, empowerment to learn, and employee development) on key department performance measures including customers' service satisfaction.

Despite our prior knowledge of how the real system operates, we have been unable to hypothesize precise differences in variable relationships, instead predicting general patterns of relationships. Indeed, ABMS is a decision-support tool and is only able to inform us about directional changes between variables (actual figures are notional).

In the broader simulation literature there is some divergence about whether or not it is appropriate to apply rigorous statistical tests in the analysis of simulation results (e.g. Schmeiser, 2001). Law and Kelton (2000) advocate the application of T-tests and not ANOVAs; nevertheless ANOVAs and T-tests are a similar type of statistical test (they are both parametric tests), and so both rely on the same key assumptions (e.g. see Howell, 2007, Pallant, 2001). Before applying any parametric test it is essential that appropriate preliminary tests check these assumptions, and where these are not met then appropriate corrections are applied. This systematic approach ensures that appropriate statistical tests are applied in the correct way.

We conduct independent replications with our simulation model, resulting in independent observations. Specifically, in line with Law & Kelton (2000) each run uses: separate sets of different random numbers (i.e. not common random numbers); the same initial conditions; and resets the statistical counters. For these reasons we are confident that we can make an assumption fundamental to the application of rigorous statistical tests, specifically the assumption of independence of observations.

### 6.1  Model Validation

To test the operation of our simulation model and ascertain confidence in the validity of our model we have designed and run two sets of validation experiments for both departments. Firstly we will look at the impact of varying the department staffing configuration on performance measures, and secondly the impact of satisfaction weights on overall customer satisfaction. All experiments hold the overall number of staffing resources constant at 10 staff and we run the simulation for a period of 10 weeks. We have conducted 20 repetitions for every experimental condition.

#### 6.1.1  Staffing Configuration

During our time in the case study departments, we observed that the number of cashiers available to serve customers would fluctuate over time. In the first investigation we vary the staffing arrangement (i.e. the number of cashiers) and examine the impact on the volume of sales transactions and two levels of customer satisfaction described earlier; number of satisfied customers and overall customer satisfaction. In reality, we observed that allocating extra cashiers would reduce the shop floor sales team numbers, and therefore the total number of customer-facing staff in each department is kept constant at 10.

*6.1.1.1 Hypotheses*

Our case study work has helped us to identify the distinguishing characteristics of the departments, for example higher customer arrival rates in WW compared to A&TV, and longer service times in A&TV compared to WW. We expect these inherent differences to impact on department performance, and we therefore predict that for each of our dependent measures: number of sales transactions (1), number of satisfied customers (2) and overall customer satisfaction (3):
- H1a, H2a, H3a: An increase in the number of cashiers will be linked to increases in 1, 2 and 3 respectively to a peak level, beyond which 1, 2 and 3 will decrease.
- H1b, H2b, H3b: The peak level of 1, 2 and 3 respectively will occur with a smaller number of cashiers in A&TV than in WW.

*6.1.1.2 Results*

Preliminary analyses were conducted for each department. The distributions of all 3 dependent variables are approximately normal (Kolmogorov-Smirnov statistics all $p>.05$). For 1 and 2, Levene's test of equality of variances was violated ($p<.05$). It has been credibly established that this is not a problem; ANOVAs are robust to violations of this assumption provided that the size of the groups are reasonably similar (Stevens, 1996), and in our case the group sizes are identical ($n=20$). Therefore it is appropriate to analyze each dependent variable using a two-way between-groups analysis of variance (ANOVA). Where significant ANOVA results were found, post-hoc tests have been applied where appropriate to investigate further the precise impact on outcome variables under different experimental conditions. To address the increased risk of a Type I error associated with multiple tests we have applied a Bonferroni correction to create more conservative thresholds for significance (corrected post-hoc $p$-value for 3 dependent variables = .0167).

Each ANOVA revealed statistically significant differences (see Table 3 for descriptive statistics). For the number of sales transactions (1) there were significant main effects for both department [$F(1, 190) = 356441.1$, $p<.001$] and staffing [$F(4, 190) = 124919.5$, $p<.001$], plus a significant moderating effect of department type [$F(4, 190) = 20496.37$, $p<.001$]. Tukey's post hoc tests were run to explore the impact of staffing and revealed significant differences for every paired comparison ($p<.001$).

[INSERT TABLE 3]

There is clear support for H1a. We expected this to happen because the number of normal staff available to provide customer advice will eventually reduce to the extent where there will be a detrimental impact on the number of customers making a purchase. Some customers will become impatient waiting increasingly long for service, and will leave the department without making a purchase. H1b is not supported, the data presents an interesting contrast, in that 1 plateaus in A&TV around 3 and 4 cashiers, whereas WW benefits greatly from the introduction of a fourth cashier. Nonetheless this finding supports the thinking underlying this hypothesis, in that we expected the longer average service times in A&TV to put a greater 'squeeze' on customer advice with even a relatively small increase in the number of cashiers.

For the number of satisfied customers (2), there were significant main effects for both department [$F(1, 190) = 391333.7$, $p<.001$], and staffing [$F(4, 190) = 38633.83$, $p<.001$], plus a significant moderating effect of department type [$F(4, 190) = 9840.07$, $p<.001$]. Post hoc tests explored the impact of staffing, and revealed significant differences for every single comparison ($p<.001$).

The results support both H2a and H2b. We interpret these findings in terms of A&TV's greater service requirement, combined with the reduced availability of advisory sales staff. These factors result in a peak in the number of satisfied customers with a smaller number of cashiers (4) than in WW (5).

For overall customer satisfaction (3), there were significant main effects for both department [$F(1, 190) = 117214.4$, $p<.001$], and staffing [$F(4, 190) = 29205.09$, $p<.001$], plus a significant moderation effect of department type [$F(4, 190) = 6715.93$, $p<.001$]. Tukey's post hoc comparisons indicated significant differences between all staffing levels ($p<.001$).

Our results support H3a for A&TV, showing a clear peak in overall customer satisfaction. H3a is only partially supported for WW, in that no decline in 3 is evident with up to 5 cashiers, although increasing this figure may well expose a peak because the overall customer satisfaction is starting to plateau out. The results offer firm support in favor of H3b.

### 6.1.2 Sensitivity of the Service Level Index

ManPraSim v1 incorporates a novel way of measuring customers' service satisfaction. It is a new feature of the model, and before we progress to investigate management practices it is important to conduct a sensitivity study with the satisfaction weights and assess the impact on the overall customer satisfaction measure.

For this series of experiments we will focus on the two main customer activities involving the interaction between customers and staff: buying and asking for help. We have switched off the refund loop because it would not add any relevant information to these results. Therefore we have two main customer blocks (pay block and help block) where we will systematically change the satisfaction weights settings to observe what effect these changes have on overall customer satisfaction. We will use the same staffing configuration for both departments; 3 cashiers, 6 normal selling staff, and 1 expert.

Our original service level index configuration allocates a satisfaction weight at each relevant customer transition (see Figure 3). Depending on the impact of a transition, the linked satisfaction weights will be either 1, 2 or 4, and can be positive or negative depending on whether or not the customer perceives the interaction positively or negatively. We will investigate 3 different scenarios each with 3 levels of satisfaction weights.

In scenario 1 each satisfaction weight has been set to the same value. Different experimental conditions have been created by systematically increasing this single value. This has changed the relative relationship between the weights from the standard model implemented described above (see Section 5.3). For scenario 2 the satisfaction weights have all been multiplied by the 3 different values for each level (1, 10, and 100). This keeps the inter-relationships between service level index values constant. In scenario 3 we have increased satisfaction weights within the same experiment in order to investigate more extreme inter-relationships between satisfaction weights. The first set are the standard satisfaction weights (see Section 5.3, 1-2-4), the second set are the first set squared (see Section 5.3, 1-4-16), and the third set are the second set squared for a further time (1-16-256). Logically we expect that increasing the satisfaction weights will result in higher overall customer satisfaction. Comparing the two departments, we expect increments in satisfaction weights to be positively linked to higher overall customer satisfaction in WW than in A&TV, given the higher visitor arrival rates and higher conversion rates in WW. The A&TV department has a higher proportion of customers requiring advice, and when the staffing levels are held constant we expect this will mean a relatively small growth in overall customer satisfaction compared to WW. For the third scenario, in A&TV customer demand for A&TV department's expert is likely to negatively impact on overall customer satisfaction. This occurs because a customer who leaves prematurely, whilst waiting for expert advice, results in the highest satisfaction penalty.

*6.1.2.1 Hypotheses*

Therefore we predict the following hypotheses:
- H4. For scenario 1, we predict a uniformly positive and linear relationship between satisfaction weights and overall customer satisfaction in both departments.
- H5. For scenario 1, we predict that increasing the satisfaction weights will have a greater positive impact in WW than A&TV.
- H6. For scenario 2, we predict a uniformly positive and linear relationship between satisfaction weights and overall customer satisfaction in both departments.
- H7. For scenario 2, we predict that increasing the satisfaction weights will have a greater positive impact in WW than A&TV.
- H8. For scenario 3, we predict a uniformly positive and non-linear relationship between satisfaction weights and overall customer satisfaction in both departments.
- H9. For scenario 3, we predict that increasing the satisfaction weights will have a greater positive impact in WW than A&TV.

*6.1.2.2 Results*

Preliminary tests confirmed that the distributions of overall customer satisfaction, for both departments, are approximately normal (Kolmogorov-Smirnov statistics all $p > .05$). For Scenarios 1, 2 and 3 Levene's test of equality of error variances was violated ($p<.01$). Therefore the significance value has been set to a stricter level ($p<.01$). Where significant ANOVA results were found, post-hoc tests have been applied where appropriate to investigate further the precise relationship between satisfaction weights and overall customer satisfaction.

Each ANOVA revealed statistically significant differences (see Table 4 for descriptive statistics). For scenario 1 there were significant main effects for both department [$F(1, 114) = 30,363.42$, $p<.01$] and satisfaction weights [$F(2, 114) = 2,943.58$, $p<.01$], plus a significant interaction effect [$F(2, 114) = 2,439.80$, $p<.01$]. The effect sizes are very large (partial eta-squared = .996, .998, and .977 respectively). Post-hoc comparisons confirmed that every single paired comparison exhibited a significant difference ($p<.01$).

[INSERT NEW TABLE 4]

Results for Scenario 1 offer support for both H4 and H5. The pattern of relationships can be clearly seen in Figure 4; overall customer satisfaction rises with the satisfaction weights in both departments, but to a greater extent in WW as confirmed by the significant interaction effect.

[INSERT FIGURE 4]

Investigating scenario 2, an ANOVA identified statistically significant main effects for department [$F(1, 114) = 32,726.63$, $p<.01$] and satisfaction weights [$F(2, 114) = 152,387.00$, $p<.01$], with a significant interaction effect [$F(2, 114) = 23,929.13$, $p<.01$]. The effect sizes are substantial (partial eta-squared = .997, 1.00, and .998 respectively). Tukey's post-hoc tests revealed significant differences ($p<.01$) for every paired comparison.

Findings support H6; the satisfaction weight is significantly related to overall customer satisfaction. Results also support H7 whereby department type is linked to significantly higher overall satisfaction values in WW than in A&TV. The pattern of relationships is graphically displayed in Figure 5.

[INSERT FIGURE 5]

Finally, for scenario 3, an ANOVA revealed statistically significant main effects for department [$F(1, 114) = 13,280.03$, $p<.01$] and satisfaction weight [$F(2, 114) = 2,771.02$, $p<.01$], with a significant interaction effect [$F(2, 114) = 8,544.38$, $p<.01$]. The effect sizes are substantial (partial eta-squared = .991, .980, and .993 respectively). Post-hoc comparisons revealed significant differences ($p<.01$) between every single paired comparison.

For scenario 3 the use of multiple satisfaction weights made it impossible to fully account for the variability of values in a single experiment. For this reason a proxy has been used; the middle value (2, 4, 16). There is partial support for H8; for WW we can see a steady increase in the overall customer satisfaction with higher squared input values. Surprisingly however we observe the inverse for A&TV; with higher squared values the overall customer satisfaction falls considerably. It appears that the greater service requirement linked to A&TV customers interacts with more extreme overall customer satisfaction scores, resulting in very negative overall customer satisfaction. Figure 6 graphically displays these relationships and suggests non-linear associations (NB the A&TV non-linear relationship is in the opposite direction). Results support H9; WW is linked to higher overall customer satisfaction than A&TV. We expected this pattern because the greater service component of A&TV roles means that when customers' demands increase, any surplus staff capacity for dealing with requests is filled and further customer requirements cannot always be satisfactorily met. In turn this triggers an increase in customer waiting times and customers are five times more likely to leave prematurely than in WW.

[INSERT FIGURE 6]

### 6.1.3 Validation Summary

Investigation of the impact of staffing configuration has provided support for most of our hypotheses. As these general patterns in the data were as we would expect, this builds our confidence in the accuracy and validity of the model. Comparing different customers' service satisfaction scenarios has demonstrated that changing the relative differences between satisfaction weightings, and not the absolute differences, has a greater impact on customers' service satisfaction measures. These validation experiments have informed the standard configuration of ManPraSim v1 which has been used in the following experiments of management practices.

The overall validation process permits us to conclude that our hypotheses have been largely supported. We are satisfied that we can have sufficient confidence in the ability of our model to provide valid results to progress and investigate more complex phenomena, specifically the impact of people management practices on performance measures.

### 6.2 Management Practice Experiments

We have designed 3 experiments to investigate the impact of task empowerment, empowerment to learn, and employee development. Global model settings for A&TV (the department under investigation in these experiments) are held constant across these experiments; the staff group in every experiment consists of 3 cashiers, 7 normal selling staff and 2 experts, with a customer arrival rate of 70 per hour and a runtime of 10 weeks. We have systematically manipulated only the independent variable of interest in each experiment. We have conducted at least 20 replications for every experimental condition enabling thorough analysis of the results.

### 6.2.1 Task Empowerment

During our case study work, we observed the implementation of a new refund policy. This new policy allows any cashier to decide independently whether to make a customer refund up to the value of £50, rather than being required to refer the authorization to an expert employee. To simulate the impact of this practice on key business outcomes, we have systematically varied the probability that employees are empowered to make refund decisions autonomously. Cashiers were configured to process a refund in 80% of cases, whereas experts were more critical and only accept 70% of refund claims.

*6.2.1.1 Hypotheses*

As we increase the level of empowerment, we expect to see more transactions as work flows more effectively and cashiers can take more decisions autonomously and quickly without requiring expert assistance. We also anticipate greater levels of overall customer satisfaction (whether obtaining a refund or not), because staff time is less consumed by the delays of locating expert assistance, resulting in more employee time available to customers. As the level of empowerment increases, we predict:
- H10. higher numbers of transactions.
- H11. higher overall customer satisfaction (shopping).
- H12. higher overall customer satisfaction (refund).

*6.2.1.2 Results*

Preliminary analyses tested the assumptions of rigorous statistical tests. The distributions of all 3 dependent variables approximate to the normal distribution (Kolmogorov-Smirnov statistics all $p>.05$). Levene's test of homogeneity of variances was violated by the number of transactions only ($p=.02$), although this does not need to be corrected for, given the equal group sizes ($n=20$, Stevens, 1996). Consequently we can proceed to apply a series of one-way between groups ANOVAs. Where significant ANOVA results were found, post-hoc tests have been applied to investigate further the precise impact on outcome variables under different experimental conditions. To address the increased risk of a Type I error we have applied a Bonferroni correction (corrected post-hoc *p*-value for 3 dependent variables = .0167).

ANOVAs revealed statistically significant differences across all three outcomes: number of transactions [$F(4, 95)=26.77$, $p<.01$], overall customer satisfaction (shopping) [$F(4, 95)=12.35$, $p<.01$], and overall customer satisfaction (refund) [$F(4, 95)=2001.73$, $p<.01$]. Consulting Table 5, we see that H10 has not been supported, and the number of transactions actually decreases with empowerment, whereas H11 and H12 are confirmed. The effect size, calculated using eta squared, reveals differences in the relative impact of empowerment on each outcome measure: 0.53 for the number of transactions, 0.34 for overall customer satisfaction (shopping) and 1.00 for overall customer satisfaction (refund). Social scientists report 0.14 as indicative of a large effect (Cohen, 1988) suggesting we are looking at substantial effect sizes.

[INSERT TABLE 5]

Post-hoc comparisons using Tukey's test indicated a number of significant differences between group means. Most notably the impact on overall customer satisfaction (refund) was great, with every single increment in empowerment resulting in a significant increase in overall customer satisfaction (refund). H10 has not been supported. This unforeseen reduction in transactions has occurred because less experienced employees take longer to make a decision on a refund, resulting in a knock-on impact for customer waiting times. H11 holds, and this finding is intuitive because customers prefer that one staff member can deal with their needs. H12 is strongly supported, and makes sense because cashiers are also more likely to approve a customer refund request.

### 6.2.2 Empowerment to Learn

Our case study work has revealed that a key way in which employees can develop their product knowledge occurs when they are unable to fully meet a customer's request for advice. The employee calls over an expert colleague and the original employee is empowered to choose whether or not to stay with them to learn from the interaction. In this second set of experiments we are assuming that, given the opportunity to choose to learn, an employee will usually decide to take up that opportunity. We found that case study employees enjoyed providing excellent customer service, and given the opportunity would do what they could to stay abreast of product developments.

In our model, a normal staff member gains knowledge points on every occasion that he or she stays with an expert to learn from a customer interaction. We have systematically varied the probability that a normal staff member learns in this way. Naturally we expect there to be a trade-off with short-term ability to meet customer demand, and a customer may leave prematurely if they have to wait for too long. A normal staff member will be occupied for longer when his or her will to learn is stronger.

*6.2.2.1 Hypotheses*

By allowing employees to acquire new product knowledge from expert colleagues, we anticipate performance improvements. We predict that increasingly empowering employees to learn will result in:
- H13. an increase in the knowledge of normal staff.
- H14. an increase in the utilization of normal staff.
- H15. no change to the utilization of expert staff.
- H16. a short term reduction in the number of sales transactions.
- H17. a reduction in overall customer satisfaction.

*6.2.2.2 Results*

Preliminary tests confirm that the distribution of all[2] 5 dependent variables are approximately normal (Kolmogorov-Smirnov statistics all $p>.05$). Levene's test of homogeneity of variances was violated by normal expertise[3] only ($p<.01$); given the near-equal group sizes ($n=20$ or 21) we can safely continue (Stevens, 1996) and it is to appropriate to use a series of one-way between groups ANOVAs. Post-hoc tests were run to follow-up significant ANOVA results, and again a Bonferroni correction was applied (corrected post-hoc *p*-value for 5 dependent variables = .01).

---
[2] Normal staff member expertise was not tested in the zero empowerment condition because it does not vary.
[3] This largely related to the fact that normal expertise is constant (=0) in the condition where the probability of an employee learning from a customer interaction equals zero.

The ANOVAs (see Table 6 for descriptives) exposed a significant impact of empowerment to learn on: normal staff expertise [$F(4, 96)=2,794.12$, $p<.01$], utilization of normal staff [$F(4, 96)=112.53$, $p<.01$], and overall customer satisfaction [$F(4, 96)=29.16$, $p<.01$]. Tests of expert staff utilization [$F(4, 96)=1.28$, $p=.29$] and sales transactions [$F(4, 96)=1.25$, $p=.30$] were insignificant. Effect sizes of significant relationships were all large (normal staff expertise = 0.99, normal staff utilization = 0.83, overall customer satisfaction = 0.55).

[INSERT TABLE 6]

Tukey's post-hoc comparisons were run for the three significant findings. Both normal staff expertise and utilization significantly increased with every single increment in employee empowerment to learn. The largest significant differences in overall customer satisfaction are observed at the polar ends of the scale. As predicted, employees who are empowered to learn become more knowledgeable (H13), leading to a more efficient utilization of employees as a whole (H14). H15 holds as expected, meaning there is no significant impact on the utilization of expert staff in terms of the time they spend engaged with customers. However we can see through the effects on other outcome measures that higher levels of learning empowerment result in better 'utilization' of experts; in other words more efficient harnessing of their knowledge. H16 has not been supported as the number of transactions does not significantly differ between experimental conditions. The short-lived reduction that we had anticipated is so negligible, it is inconsequential. Nonetheless, the associated increase in customer waiting times has negatively influenced the customer service index, in support of H17.

### 6.2.3 Employee development

Our final investigation of management practices goes one step further and explores how time invested in learning impacts on medium-term system performance. Our model mimics an evolutionary process whereby staff members can progressively develop their product knowledge over a period of time. When a staff member has accumulated a certain number of knowledge points from observing expert service transactions, they are considered an expert. We have systematically varied the number of knowledge points required to attain expert-level competence. All normal staff members are programmed to take advantage of all learning opportunities.

*6.2.3.1 Hypotheses*

By investing time in developing and expanding employees' specialist knowledge, we anticipate even greater future savings in terms of key outcomes, beyond those already observed in the previous experiment. The academic literature echoes the positive business impact of employing individuals with greater expertise to provide better customer service and advice (e.g. Crosby et al., 1990). We predict that increasing the rate of employee development (by lowering the threshold for attaining expert status) will result in more desirable outcome variables, specifically increases in:
- H18. normal staff member expertise.

- H19. normal staff utilization.
- H20. expert staff utilization.
- H21. the number of transactions.
- H22. overall customer satisfaction.

*6.2.3.2 Results*

Preliminary tests confirmed the distribution of all[4] 5 dependent variables are approximately normal, with the exception of expert staff utilization for a promotion criteria level of 0.8 (Kolmogorov-Smirnov statistic $p<.01$). This non-normal distribution will have minimal impact on the significance or power of the test (Stevens, 1996, p.240). Levene's test of homogeneity of variances was violated by all 5 dependent variables[5] ($p<.05$); again we rely on Steven's (1996) empirical argument that because we have equal group sizes ($n=20$) this is acceptable. We have applied a series of one-way between groups ANOVAs with Tukey's post-hoc tests to determine the specific nature of significant ANOVA results, and again a Bonferroni correction was applied (corrected post-hoc *p*-value for 5 dependent variables = .01).

The final ANOVAs revealed statistically significant differences in expert utilization [$F(5,114)=952.21$, $p<.01$], volume of transactions [$F(5,114)=193.14$, $p<.01$] and overall customer satisfaction [$F(5,114)=959.01$, $p<.01$]. The effect sizes of significant relationships were again all very large (expert staff utilization = 0.98, volume of transactions = 0.89, and overall customer satisfaction = 0.98). We were unable to adequately test the impact of learning on normal staff expertise (H18) and utilization (H19), because we do not have this data for all experimental conditions (see Table 7: at the lower promotion thresholds, all normal staff have been promoted before the end of the simulation run and so these values are recorded as zero).

[INSERT TABLE 7]

Tukey's post-hoc comparisons revealed significant differences between every single increment in the competence threshold for each variable, with the exception of the two upper levels. However, only expert utilization was in the predicted direction. Our evidence was strongly in favor of H20, whereas the exactly the contrary of H21 and H22 have been supported. This is counter-intuitive because we had expected that the greater the number of resulting experts, the greater the availability of top-quality advice to customers. Indeed, it is possible that our simulation run is too short at just ten weeks, and presents only a backward facing view of department performance; i.e., focusing on the time consumed in learning, and not on the time spent sharing their new competence with customers. If we ran the simulation for longer we would still expect our original hypotheses to hold true. In our model we are also assuming that staff acquire expertise purely by learning from their colleagues, whereas in reality this would be supported with other sources and forms of learning.

---

[4] The distributions of normal staff expertise and normal staff utilization have not been tested for promotion criteria levels 0, 0.2, 0.4, and 0.6 because these variables are constant (=0) under these conditions.

[5] The distributions of normal staff expertise and normal staff utilization have not been tested for promotion criteria levels 0, 0.2, 0.4, and 0.6 because these variables are constant (=0) under these conditions.

### 6.2.4 Summary and Implications of Management Practice Experiments

All in all we can conclude that through modeling and simulating the impact of management practices we are able to evaluate hypotheses regarding department performance. Although some empowerment effects may at first appear small or even the inverse of what we had expected, over time these changes can have a meaningful impact. For this reason it is important to incorporate management practices in a retail department model. It is also worth noting that the full benefits of empowering staff may also be in terms which our simulator does not currently measure, such as an employee's satisfaction with their job or their intention to stay with the business. It is therefore likely that our performance figures are an under-estimate of the true impact of the various ways of empowering shop-floor staff that we have looked at.

From our experimental investigations, it is apparent that some management practices offer greater potential for performance improvements than others. Our results lead us to make a number of suggestions for the A&TV department to which they pertain. The first experiment investigated the impact of empowering staff to make refunds and demonstrated improvements in overall customer satisfaction (shopping and refund), but the number of transactions did not increase as we had expected. Employees who are less experienced at processing refunds tend to take longer than someone who has more experience. The implication for retailers is to ensure that each employee receives sufficient training, and has the opportunity to fully familiarize himself or herself with the refund process. Carefully implementing such a scheme would minimize the amount of time these employees need to get fully 'up to speed' with their new responsibilities on the shop-floor.

The second experiment examined the influence of employees being empowered to learn new product knowledge from expert colleagues advising customers on the shop-floor. Our results suggest that there are performance improvements to be gained through encouraging expert employees to share their knowledge with less-experienced colleagues 'on-the-job'. Retailers need to be in it for the long-term, in that there is a short-term trade off with customers' service satisfaction whilst less experienced staff develop new capabilities. In reality an employer would benefit from aligning this kind of scheme with a staff retention initiative to ensure that workers are committed to staying with the organization and contributing to its long-term success.

The third and final experiment looked at the impact of employee development and how time invested in learning impacts on medium-term system performance. Our findings indicated that the time it takes to see performance benefits is longer than 10 weeks. We can, nevertheless, advise that to establish the optimal level of employee training in a given context, the particulars of that situation would first need to be more closely examined. For example, the cost of employee wages, availability of skilled labor, and customers' expectations of shop-floor staff can all vary to some extent by geography and the specific nature of the work. Consequently this example does not allow us to conclude more than illustrating the point that employee development takes time to positively impact on performance.

## 7     CONCLUSIONS AND FUTURE DIRECTIONS

In this paper we have described our first approach to modeling people management practices, in particular employee development and empowerment. We have also described the development of a novel performance measure to gauge customers' service satisfaction. In this version of the simulation model we use customer agents that are fairly homogeneous and have no enduring memory (they only visit the shop once in their lifetime). Still, with this simulation model we have been able to demonstrate how one could implement different empowerment management practices in a simulation model to investigate their effect on customers' service satisfaction, the development of staff knowledge and sales turnover.

As far as we are aware, this is the first time that researchers have tried to use an agent-based approach to simulate management practices such as employee development and empowerment. Although our simulation model has been driven by case studies from one retail organization, we believe that the general model could be adapted to other businesses operating in different industries and countries, for businesses which use management practices involving a high degree of human interaction.

From what we can conclude from our current analyses, some findings are as we hypothesized whereas others are more mixed. Two sets of validation experiments provide us with confidence in the model's ability to produce reliable and valid results. Early findings indicate that management practices tend to exert a subtle yet significant effect on performance, consistent with our case study findings. Further experimentation is required to explore the model's operation, and more development work would bring the model closer to the real system and allow us to incorporate more complex and interesting features.

We have identified two major limitations that we want to address in our future work. Firstly, the current version of the simulation model has a high level of abstraction which represents the actors of the real system and their behaviors in a simplistic way, cutting out some relevant features and behaviors. For example, the homogeneity of our customers prevents us from representing the true variability of customer perceptions of, and responses to, service. Currently the model represents customer perceptions and responses to the same situation in the same way. Secondly, in the current version we cannot measure any long-term performance effects because we have an infinite number of homogenous customers who only visit the department once. It is important to overcome this limitation because many organizations strive to retain their customers over time.

In the second version of the simulation model we want to address the shortcomings we have identified above. Our main focus will be to make the simulation a better representation of the real system (i.e. the retail department) and the people in the real system, including their characteristics and behavior. We will incorporate realistic footfall data, heterogeneous customer types, a finite population in the form of a customer pool, a prompt departure of customers at closing time, and an enhanced version of our novel customer service satisfaction measure.

Adding customer types and thereby diversifying the population will allow us to represent varied perceptions of and reactions to the shopping experience. Our case study organization has identified the particular customer types which are important to its business through market research, and we plan to find out how populations of certain customer types influence sales. Through introducing a finite population each customer agent will have a memory of his or her previous shopping experiences which will impact on how the agent perceives the customer service on his or her next visit. As we observed in Section 4.2.4 a bad experience might make him or her come less frequently or might reduce subsequent patience level thereby reducing the probability of that customer buying something.

Taking a step back, we believe that researchers need to become more involved in this multi-disciplinary kind of work to gain new insights into the behavior of organizations. In our view, the main benefit from adopting this approach is the improved understanding of and debate about a problem domain, and the resulting explicit convergence of understanding and agreement about a system's functioning. The very nature of the methods involved forces researchers to be explicit about the rules underlying behavior and to think in new ways about them. As a result, we have brought work psychology and ABM closer together to develop a new and exciting research area.

# FIGURES AND TABLES

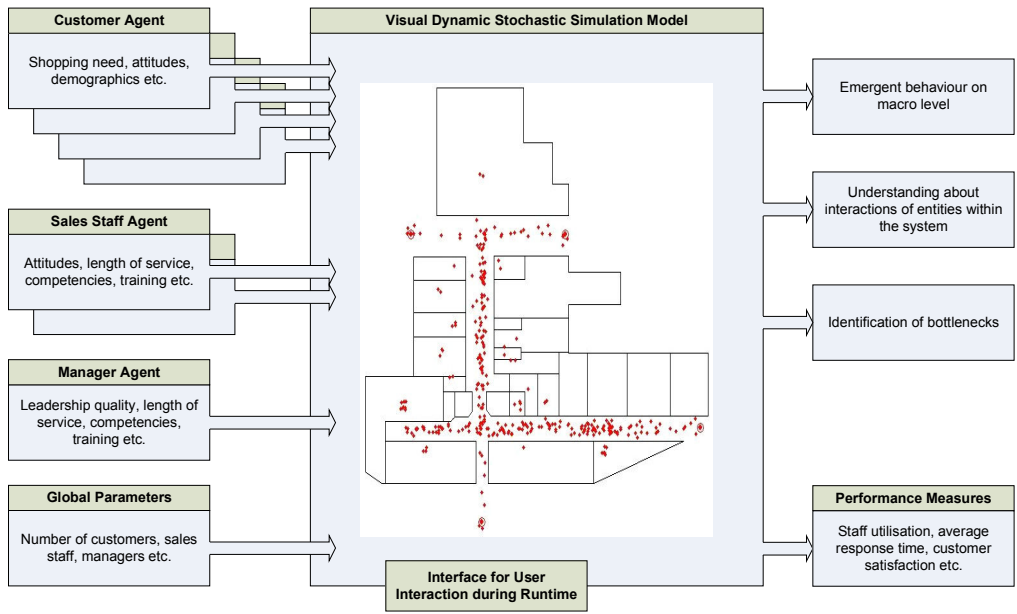

Figure 1. Conceptual model of the simulation model.

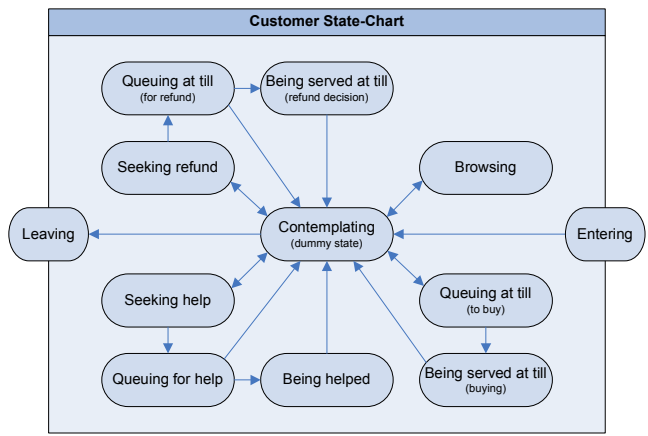

Figure 2. Conceptual model of a customer agent.

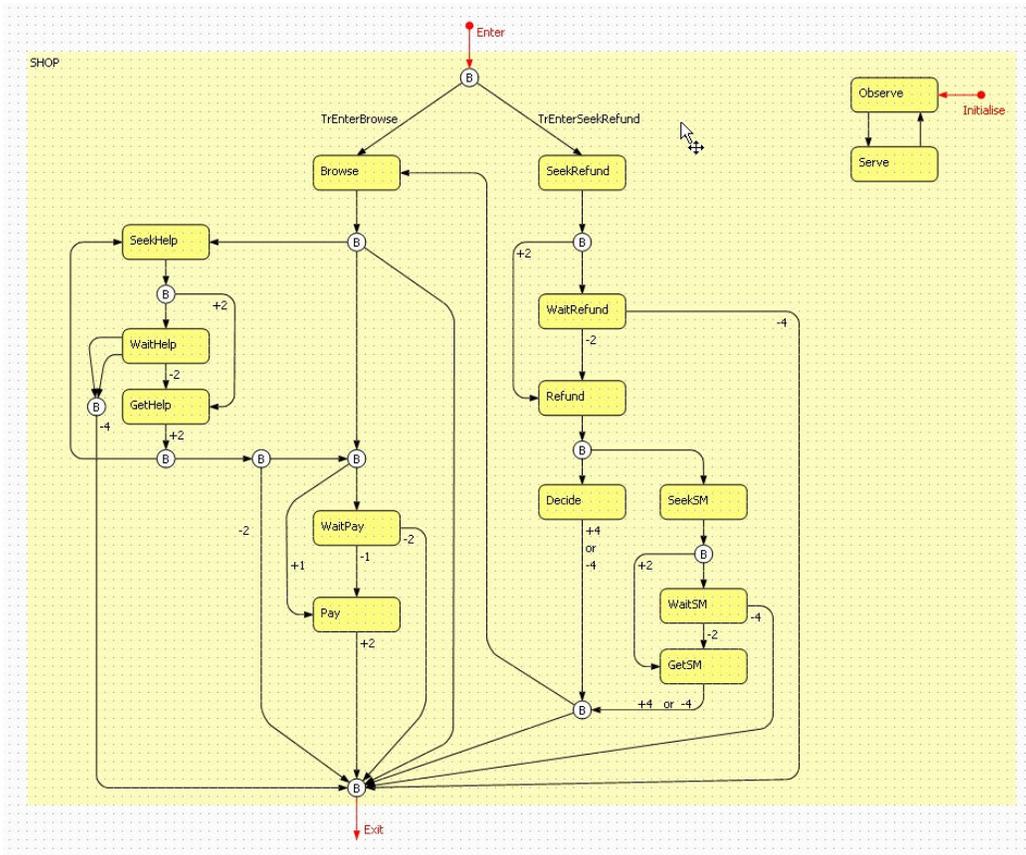

Figure 3. Customer (left) and staff (right) agent logic implementation in AnyLogic™.

| Situation | Min | Mode | Max |
|---|---|---|---|
| leave browse state after … | 1 | 7 | 15 |
| leave help state after … | 3 | 15 | 30 |
| leave pay queue (no patience) after … | 5 | 12 | 20 |

Table 1. Sample frequency distribution parameter values.

| Event | Probability of event |
|---|---|
| someone makes a purchase after browsing | 0.37 |
| someone requires help | 0.38 |
| someone makes a purchase after getting help | 0.56 |

Table 2. Sample probabilities.

| Department | Cashiers | Number of transactions | | Number of satisfied customers | | Overall customer satisfaction | |
|---|---|---|---|---|---|---|---|
| | | Mean | S.D. | Mean | S.D. | Mean | S.D. |
| A&TV | 1 | 4853.50 | 26.38 | 12324.05 | 77.64 | 9366.40 | 563.88 |
| | 2 | 9822.20 | 57.89 | 14762.45 | 81.04 | 19985.20 | 538.30 |
| | 3 | 14279.90 | 96.34 | 17429.70 | 103.77 | 28994.80 | 552.60 |
| | 4 | 14630.60 | 86.19 | 17185.00 | 99.09 | 32573.60 | 702.64 |
| | 5 | 13771.85 | 97.06 | 16023.20 | 82.66 | 27916.05 | 574.56 |
| WW | 1 | 8133.75 | 22.16 | 18508.20 | 88.68 | 17327.95 | 556.03 |
| | 2 | 15810.10 | 56.16 | 22640.40 | 92.00 | 42339.10 | 736.61 |
| | 3 | 25439.60 | 113.66 | 28833.10 | 115.65 | 58601.10 | 629.68 |
| | 4 | 30300.70 | 249.30 | 32124.60 | 230.13 | 74233.30 | 570.79 |
| | 5 | 28894.25 | 195.75 | 30475.20 | 176.41 | 76838.65 | 744.31 |

Table 3. Descriptive statistics for the first validation experiment (to 2 d.p.).

| Department | Satisfaction weight value level | Scenario 1 | | Scenario 2 | | Scenario 3 | |
|---|---|---|---|---|---|---|---|
| | | Mean | S.D. | Mean | S.D. | Mean | S.D. |
| A&TV | 1 | 16,381.20 | 242.42 | 27,624.55 | 797.73 | 27,624.55 | 797.73 |
| | 2 | 33,095.00 | 568.22 | 274,347.00 | 4,061.43 | 27,541.75 | 1,973.94 |
| | 3 | 49,720.20 | 881.91 | 2,743,685.00 | 59,264.25 | -581,334.95 | 29,429.71 |
| WW | 1 | 30,383.40 | 286.14 | 63,219.15 | 812.36 | 63,219.15 | 812.36 |
| | 2 | 60,507.80 | 699.92 | 632,188.00 | 8,101.63 | 121,107.70 | 1,780.45 |
| | 3 | 90,687.60 | 1,655.07 | 6,344,035.00 | 78,434.26 | 260,411.75 | 23,349.43 |

Table 4. Descriptive statistics for the second validation experiment (to 2 d.p.).

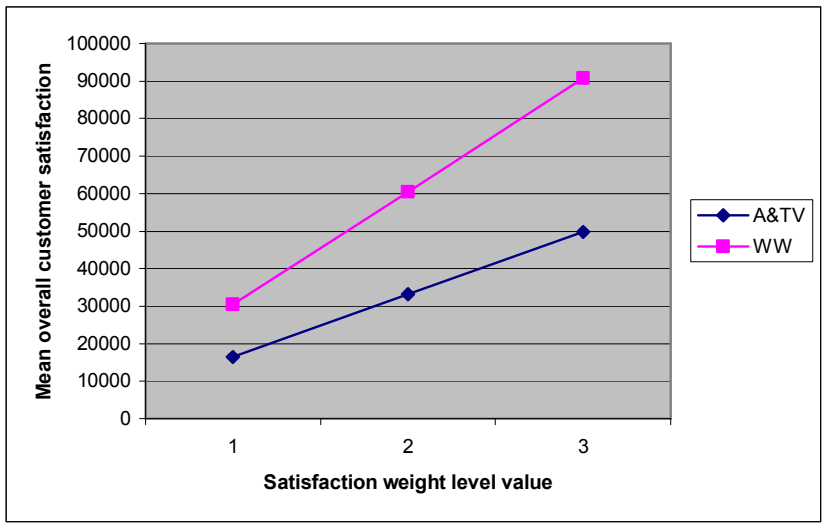

Figure 4. Scenario 1: Satisfaction weight level value by mean overall customer satisfaction.

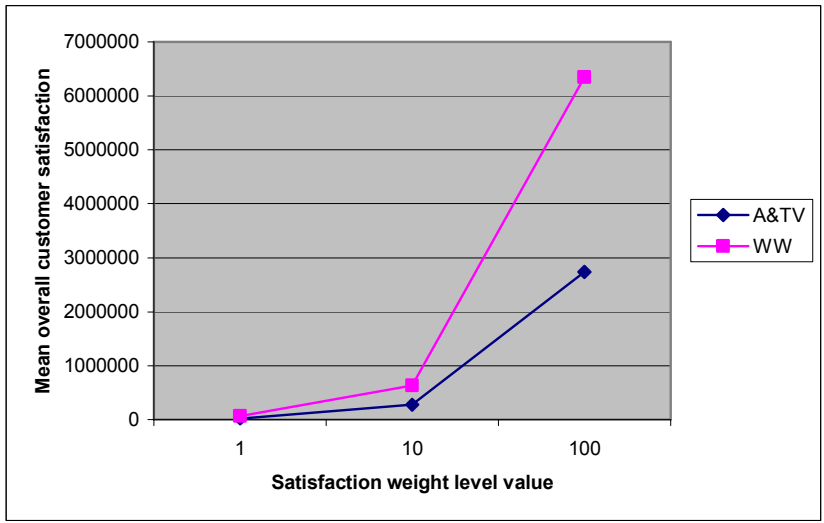

Figure 5. Scenario 2: Satisfaction weight level value by mean overall customer satisfaction.

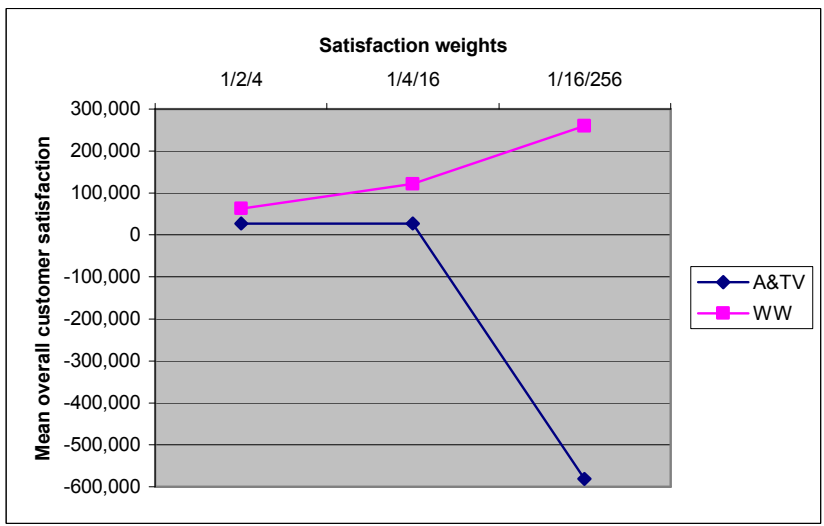

Figure 6. Scenario 3: Satisfaction weights by mean overall customer satisfaction.

| Empower-ment level | Number of transactions | | Overall customer satisfaction for shopping | | Overall customer satisfaction for refund | |
|---|---|---|---|---|---|---|
| | Mean | S.D. | Mean | S.D. | Mean | S.D. |
| 0 | 15133.85 | 102.02 | 23554.30 | 892.55 | -3951.40 | 288.84 |
| 0.25 | 15114.75 | 60.04 | 24331.35 | 907.02 | -2316.10 | 187.23 |
| 0.5 | 15078.95 | 86.24 | 24476.95 | 907.48 | -932.40 | 243.25 |
| 0.75 | 15008.45 | 52.53 | 25213.10 | 898.61 | 613.70 | 182.03 |
| 1 | 14920.15 | 66.42 | 25398.95 | 1092.50 | 1892.80 | 237.69 |

Table 5. Descriptive statistics for task empowerment experiment (to 2 d.p.).

| Empower-ment level | Normal expertise | | Utilization of normal staff | | Utilization of expert staff | | Number of transactions | | Overall customer satisfaction | |
|---|---|---|---|---|---|---|---|---|---|---|
| | Mean | S.D. | Mean | S.D. | Mean | S.D. | Mean | S.D. | Mean | S.D. |
| 0 | 0.00 | 0.00 | 0.82 | 0.01 | 0.93 | 0.00 | 14830.00 | 99.82 | 28004.00 | 823.19 |
| 0.25 | 18.36 | 2.10 | 0.83 | 0.01 | 0.94 | 0.00 | 14801.00 | 73.56 | 26937.00 | 960.37 |
| 0.5 | 35.66 | 2.54 | 0.84 | 0.01 | 0.94 | 0.00 | 14782.00 | 79.90 | 26310.00 | 916.38 |
| 0.75 | 53.44 | 2.98 | 0.85 | 0.01 | 0.94 | 0.01 | 14787.00 | 96.45 | 25678.00 | 1269.68 |
| 1 | 69.35 | 2.85 | 0.85 | 0.01 | 0.94 | 0.00 | 14823.00 | 80.42 | 24831.00 | 1043.79 |

Table 6. Descriptive statistics for empowerment to learn experiment (to 2 d.p.).

| Promotional threshold | Normal staff expertise | | Normal staff utilisation | | Expert staff utilisation | | Number of transactions | | Overall customer satisfaction | |
|---|---|---|---|---|---|---|---|---|---|---|
| | Mean | S.D. | Mean | S.D. | Mean | S.D. | Mean | S.D. | Mean | S.D. |
| 0 | 0.00 | 0.00 | 0.00 | 0.00 | 0.77 | 0.01 | 15482.35 | 97.66 | 46125.25 | 1099.48 |
| 0.2 | 0.00 | 0.00 | 0.00 | 0.00 | 0.80 | 0.01 | 15302.85 | 75.00 | 40723.95 | 1209.39 |
| 0.4 | 0.00 | 0.00 | 0.00 | 0.00 | 0.82 | 0.01 | 15125.15 | 52.03 | 34992.75 | 1770.02 |
| 0.6 | 0.00 | 0.00 | 0.00 | 0.00 | 0.86 | 0.01 | 14945.30 | 118.41 | 28958.80 | 1460.78 |
| 0.8 | 67.68 | 3.23 | 0.86 | 0.01 | 0.93 | 0.02 | 14801.95 | 92.79 | 24661.75 | 1058.27 |
| 1 | 68.83 | 3.84 | 0.86 | 0.00 | 0.94 | 0.00 | 14827.90 | 76.14 | 24668.80 | 843.84 |

Table 7. Descriptive statistics for employee development experiment (to 2 d.p.).